\documentclass[letterpaper, 10 pt, conference]{ieeeconf}  
\IEEEoverridecommandlockouts                              

\usepackage{graphicx} 
\usepackage{wrapfig}
\usepackage{amsmath}
\usepackage{amsfonts}
\usepackage{amssymb}
\usepackage{bbm}
\usepackage{xcolor}

\usepackage{url}
\usepackage{cite}

\usepackage{dsfont}
\usepackage{braket}
\usepackage{booktabs}
\usepackage{color}
\usepackage{colortbl}
\usepackage{mathtools}
\usepackage{multirow}
\usepackage{footmisc}

\title{\LARGE \textbf{
Dreaming: Model-based Reinforcement Learning \\ by Latent Imagination without Reconstruction
}}

\author{Masashi Okada$^{1,\star}$ and Tadahiro Taniguchi$^{1,2}$
    \thanks{
        $^{1}$ Masashi Okada and Tadahiro Taniguchi are with Digital \& AI Technology Center, Technology Division, Panasonic Corporation, Japan.
    }%
    \thanks{
        $^{2}$ Tadahiro Taniguchi is also with Ritsumeikan University, College of Information Science and Engineering, Japan.
    }%
    \thanks{
        $^{\star}$ \texttt{okada.masashi001@jp.panasonic.com}
    }
}

\newcommand{\obs}{\boldsymbol{x}}
\newcommand{\action}{\boldsymbol{a}}

\newcommand{\latent}{{\boldsymbol z}}
\newcommand{\latents}{{\boldsymbol s}}
\newcommand{\latenth}{{\boldsymbol h}}

\newcommand{\pjoint}{p}
\newcommand{\minibatch}{\mathcal{D}}

\newcommand{\nceloss}{\mathcal{J}^{\mathrm{NCE}}}
\newcommand{\klloss}{\mathcal{J}^{\mathrm{KL}}}
\newcommand{\best}[1]{\underline{\textbf{#1}}}
\newcommand{\second}[1]{\underline{#1}}

\newcommand{\proposedmethod}{Dreaming}{}

\begin{document}

\maketitle
\thispagestyle{empty}
\pagestyle{empty}

\begin{abstract}
In the present paper, we propose a decoder-free extension of Dreamer, a leading model-based reinforcement learning (MBRL) method from pixels.
Dreamer is a sample- and cost-efficient solution to robot learning,
as it is used to train latent state-space models based on a variational autoencoder and to conduct policy optimization by latent trajectory imagination.
However, this autoencoding based approach often causes \textit{object vanishing},
in which the autoencoder fails to perceives key objects for solving control tasks, and thus significantly limiting Dreamer's potential.
This work aims to relieve this Dreamer's bottleneck and enhance its performance by means of removing the decoder.
For this purpose, we firstly derive a likelihood-free and InfoMax objective of contrastive learning from the evidence lower bound of Dreamer.
Secondly, we incorporate two components, (i) independent linear dynamics and (ii) the random crop data augmentation, to the learning scheme so as to improve the training performance.
In comparison to Dreamer and other recent model-free reinforcement learning methods,
our newly devised \textit{\underline{Dream}er with \underline{In}foMax and without \underline{g}enerative decoder} (\proposedmethod{}) achieves the best scores on 5 difficult simulated robotics tasks,
in which Dreamer suffers from \textit{object vanishing}.
\end{abstract}

\section{Introduction} \label{sec:intro}
In the present paper, we focus on model-based reinforcement learning (MBRL) from pixels without complex reconstruction.
MBRL is a promising technique to build controllers in a sample efficient manner,
which trains forward dynamics models to predict future states and rewards for the purpose of planning and/or policy optimization.
The recent study of MBRL in fully-observable environments~\cite{chua2018deep,okada2019variational,langlois2019benchmarking,kaiser2019model} have achieved both sample efficiency and competitive performance with the state-of-the-art model-free reinforcement learning (MFRL) methods like soft-actor-critic (SAC)~\cite{haarnoja2018soft}.
Although real-robot learning has been achieved with fully-observable MBRL ~\cite{bechtle2020curious,nagabandi2020deep,yang2020data,zhang2019asynchronous,williams2020locally,fang2019dynamics},
there has been an increasing demand for robot learning in partially observable environments in which only incomplete information (especially vision) is available.
MBRL from pixels can be realized by introducing deep generative models based on autoencoding variational Bayes~\cite{kingma2013auto}. 

\begin{figure}[t]
    \centering
    \includegraphics[width=0.48\textwidth]{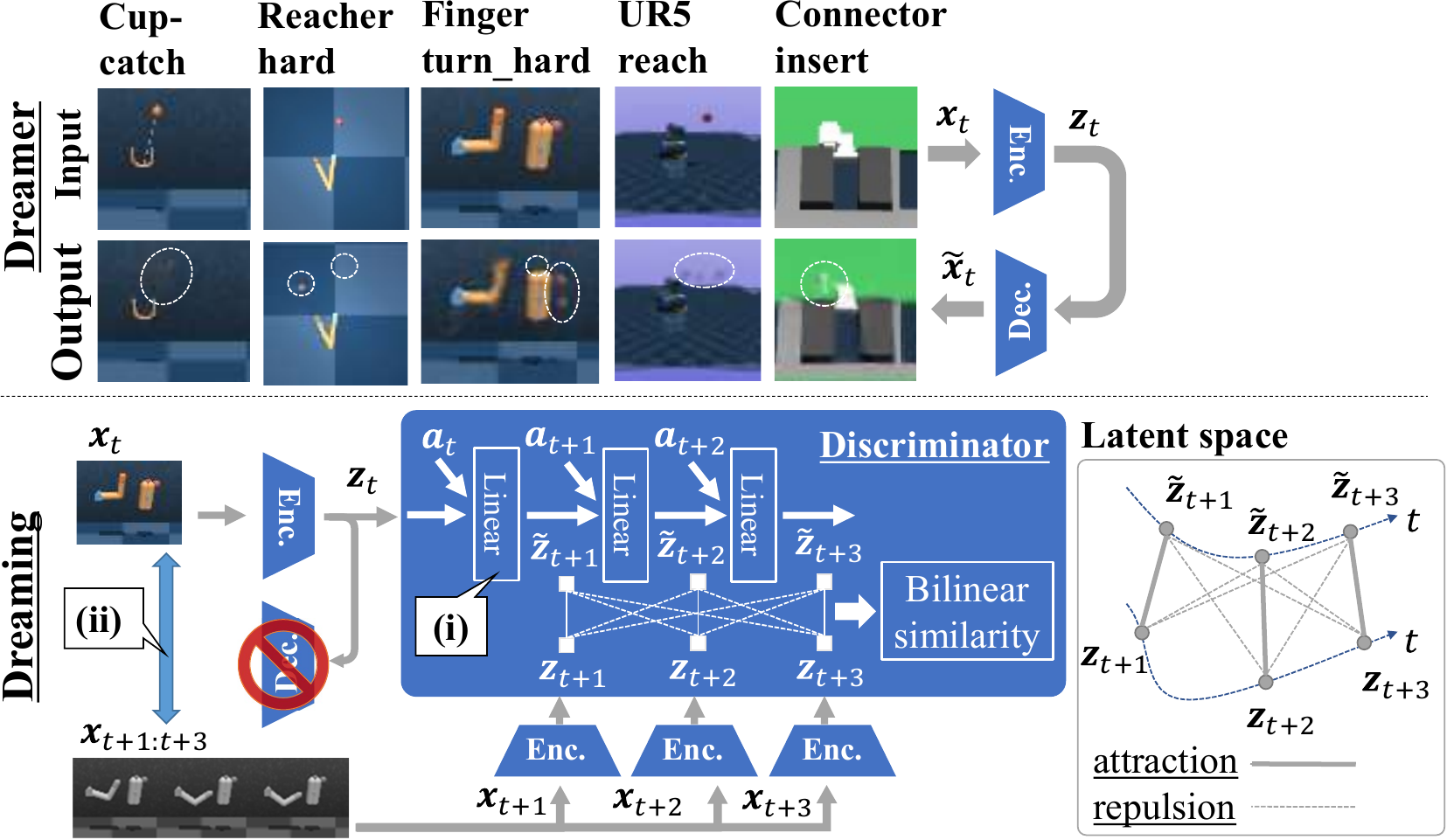}
    \caption{
    The overview of the motivation and concept of this work.
    \textbf{(Top)}
    The concept of Dreamer's autoencoding-based representation learning~\cite{hafner2019dreamer},
    which often causes \textit{object vanishing} as shown with the dashed-line ovals.
    The left three tasks are from DeepMind Control Suite ~\cite{deepmindcontrolsuite2018},
    and the remaining two tasks are our original tasks which are assumed to represent industrial applications.
    \textbf{(Bottom)}
    The concept of \proposedmethod{}'s representation learning, which trains a discriminator instead of the decoder
    to embed different samples to be spaced apart from each other.
    The learning scheme is characterized with the two key components;
    (i) Linear dynamics successfully constrains temporally consecutive samples are not distributed too further away.
    (ii) Image augmentation encourages only key features for control to be embedded into the latent space.
    }
    \label{fig:dazzling}
\end{figure}

\textit{Object vanishing} is a critical problem of the autoencoding based MBRL from pixels.
Previous studies in this field \cite{ha2018recurrent,lee2019stochastic,hafner2018learning,hafner2019dreamer,han2019variational,yarats2019improving,okada2020planet} train autoencoding models along with latent dynamics models that generate \textit{imagined} trajectories that are used for planning or policy optimization.
However, the autoencoder often fails to perceive small objects in pixel space.
The top part of Fig.~\ref{fig:dazzling} exemplifies this kind of failures where small (or thin) and important objects are not reconstructed in their correct positions.
This shows the failure to successfully embed their information into the latent space, which significantly limits the training performance.
This problem is a result of a log-likelihood objective (reconstruction loss) defined in the pixel space.
Since the reconstruction errors of small objects in the pixel space are insignificant compared to the errors of other objects and uninformative textures that occupy most parts of the image region, it is hard to train the encoder to perceive the small objects from the weak error signals.
Also, we have to train the decoder which requires a high model capacity with massive parameters from the convolutional neural networks (CNN), although the trained models are not exploited both in the planning and policy optimization.

To avoid this complex reconstruction, some previous MBRL studies have proposed decoder-free representation learning~\cite{hafner2019dreamer,yan2020learning} based on contrastive learning~\cite{oord2018representation,chen2020simple}, which trains a discriminator instead of the decoder.
The discriminator is trained by categorical cross-entropy optimization, which encourages latent embeddings to be sufficiently distinguishable among different embeddings.
Nevertheless, to the best of our knowledge, no MBRL methods have achieved the state-of-the-art results on
the difficult benchmark tasks of DeepMind Control Suite (DMC)~\cite{deepmindcontrolsuite2018} without reconstruction.

Motivated by these observations, this paper aims to achieve the state-of-the-art results with MBRL from pixels without reconstruction.
This paper mainly focuses on the latest autoencoding-based MBRL method Dreamer, considering the success of a variety of control tasks (i.e., DMC and Atari Games~\cite{bellemare2013arcade}),
and tries to extend this method to be a decoder-free fashion.
We adopt Dreamer's policy optimization without any form of modification.
We call our extended Dreamer as \textit{\underline{Dream}er with \underline{In}foMax and without \underline{g}enerative decoder} (\proposedmethod{}).
The concept of this proposed method is illustrated in the bottom part of Fig.~\ref{fig:dazzling}.
Our primary contributions are summarized as follows.
\begin{itemize}
    \item We derive a likelihood-free (decoder-free) and InfoMax objective for contrastive learning by reformulating the variational evidence lower bound (ELBO) of the graphical model of the partially observable Markov decision process.
    \item We show that two key components,
    (i) an independent and linear forward dynamics, which is only utilized for contrastive learning, and
    (ii) appropriate data augmentation (i.e., random crop),
    are indispensable to achieve the state-of-the-art results.
\end{itemize}
In comparison to Dreamer and the recent cutting edge MFRL methods, Dreaming can achieve the state-of-the-art results on difficult simulated robotics tasks exhibited in Fig.~\ref{fig:dazzling} in which Dreamer suffers from \textit{object vanishing}.
The remainder of this paper is organized as follows.
In Sec.~\ref{sec:related_work}, key differences from related work are discussed.
In Sec.~\ref{sec:prelimianary}, we provide a brief review of Dreamer and contrastive learning.
In Sec.~\ref{sec:proposed_method}, we first describe the proposed contrastive learning scheme in detail, and then introduce \proposedmethod{}.
In Sec.~\ref{sec:experiments}, the effectiveness of \proposedmethod{} is demonstrated through simulated evaluations.
Finally, Sec.~\ref{sec:conclusion} concludes this paper.

\section{Related Work} \label{sec:related_work}
\textbf{Some of the most related work} are contrastive predictive coding (CPC)~\cite{oord2018representation} and
contrastive forward model (CFM)~\cite{yan2020learning}.
Our work is highly inspired by CPC, and our contrastive learning scheme has similar components with CPC; e.g., a recurrent neural network and a bilinear similarity function.
However, \textit{CPC has no action-conditioned dynamics models}.
Since CPC alone cannot generate \textit{imagined} trajectories from arbitrary actions,
CPC is only used as an auxiliary objective of MFRL.
CFM heuristically introduces a similar decoder-free objective function like ours. 
A primary difference between ours and CFM is that \textit{CFM exploits a shared and non-linear forward model} for both contrastive and behavior learning.
In addition, {the relation between the ELBO of time-series variational inference is not discussed} in the above two literature.
Meanwhile, the original Dreamer paper \cite{hafner2019dreamer} has also derived a contrastive objective from the ELBO.
However, \textit{dynamics models and temporal correlation of observations are not involved in the contrastive objective}.
Furthermore, \textit{CPC, CFM, and Dreamer do not introduce data augmentation}.

\textbf{MFRL methods with representation learning}:
A state-of-the-art MFRL method, contrastive unsupervised representation for reinforcement learning (CURL)~\cite{srinivas2020curl}, also makes use of contrastive learning with the random crop data augmentation.
Deep bisimulation for control (DBC)~\cite{zhang2020learning} and
discriminative particle filter reinforcement learning (DPFRL)~\cite{ma2020discriminative} are other types of cutting edge MFRL methods, which utilize different concepts of representation learning without reconstruction.

\textbf{MFRL methods without representation learning}:
Recently proposed MFRL methods, which include reinforcement learning with augmented data (RAD)~\cite{laskin2020reinforcement}, data-regularized Q (DrQ)~\cite{kostrikov2020image}, and simple unified framework for reinforcement learning using ensembles (SUNRISE)~\cite{lee2020sunrise}, have achieved state-of-the-art result without representation learning.
All these work employ the random crop data augmentation as an important component of their method.

%
\begin{figure}
  \centering
    \centering
    \includegraphics[width=0.2\textwidth]{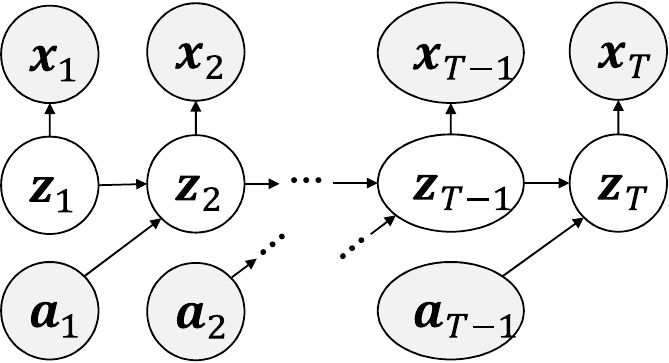}
    \caption{Graphical model of the partially observable Markov decision process.}
    \label{fig:pomdp}
\end{figure}
\section{Preliminary} \label{sec:prelimianary}
\subsection{Autoencoding Variational Bayes for Time-series} \label{sec:autoencoding}
Let us begin by considering the graphical model illustrated in Fig.~\ref{fig:pomdp},
whose joint distribution is defined as follows:
\begin{align}
  \pjoint(\latent_{\leq T}, \action_{<T}, \obs_{\leq T}) = 
    \prod_{t} p\left(\latent_{t+1} | \latent_{t}, \action_{t} \right) p\left(\obs_{t} | \latent_{t}\right),
    %
    %
\end{align}
where $\latent$, $\obs$, and $\action$ denote latent state, observation, and action, respectively.
As in the case of well-known variational autoencoders (VAEs)~\cite{kingma2013auto}, generative models $p(\latent_{t+1}|\latent_{t}, \action_{t})$, $p(\obs_{t}|\latent_{t})$ and inference model $q(\latent_{t}|\obs_{\leq t}, \action_{< t})$ can be trained by maximizing the
evidence lower bound~\cite{hafner2018learning}:
\begin{align}
  & \log p(\obs_{\leq T} | \action_{\leq T}) = \log \int \pjoint(\latent_{\leq T}, \action_{<T}, \obs_{\leq T}) d\latent_{\leq t} \geq \nonumber\\
  & \sum_{t}\left(
  \underbrace{
  \mathbb{E}_{q(\latent_{t}|\obs_{\leq t}, \action_{< t})}\left[
    \log p(\obs_{t}|\latent_{t})
  \right]
  }_{\coloneqq \mathcal{J}^{\mathrm{likelihood}}}
  \right.
  \label{eqn:vae_elbo}\\
  & \left. 
  \underbrace{
  - \mathbb{E}_{q(\latent_{t}|\cdot)}\left[
    {\operatorname{KL}}\left[
      q(\latent_{t+1} |\obs_{\leq t+1}, \action_{< t+1}) ||
      p(\latent_{t+1} |\latent_{t}, \action_{t})
  \right]
  \right]
  }_{\coloneqq \klloss}
  \right). \nonumber
\end{align}
%
If the models are defined to be differentiable and trainable, this objective can be maximized by the stochastic gradient ascent via backpropagation.

Multi-step variational inference is proposed in \cite{hafner2018learning} to improve long-term predictions.
This inference, named \textit{latent overshooting}, involves the multi-step objective $\klloss_{k}$ defined as:
\begin{align}
  \klloss \geq \klloss_{k} \coloneqq \mathbb{E}_{p(\latent_{t}|\latent_{t-k},\action_{<t})q(\latent_{t-k}|\obs_{\leq t-k}, \action_{< t-k})}[ \hookleftarrow \nonumber\\
    {\operatorname{KL}}\left[
      q(\latent_{t+1} |\obs_{\leq t+1}, \action_{< t+1}) ||
      p(\latent_{t+1} |\latent_{t}, \action_{t})\right]],
\end{align}
where
\begin{align}
    {p}(\latent_{t} | \latent_{t-k}, \action_{<t})
    \coloneqq
    \mathbb{E}_{{p}(\latent_{t-1} | \latent_{t-k}, \action_{<t-1})}
    \left[
      {p}(\latent_{t}|\latent_{t-1},\action_{t-1})
    \right] \nonumber
\end{align}
is the multi-step prediction model.

For the purpose of planning or policy optimization, not only for the dynamics model $p(\latent_{t+1}|\latent_{t}, \action_{t})$ but also the reward function $p(r_{t}|\latent_{t})$ is also required.
To do this, we can simply regard the rewards as observations and learn the reward function $p(r_{t}|\latent_{t})$ along with the decoder $p(\obs_{t}|\latent_{t})$.
For readability, we omit the specifications of the reward function $p(r_{t}|\latent_{t})$ in the following discussion.
Although we remove the decoder later,
the reward function and its likelihood objective are kept untouched.

\subsection{Recurrent State Space Model and Dreamer} \label{sec:rssm_dreamer}
The recurrent state space model (RSSM) is a latent dynamics model equipped with an expressive recurrent neural network,
realizing accurate long-term prediction.
RSSM is used as an essential component of various MBRL methods from pixels~\cite{hafner2018learning,hafner2019dreamer,han2019variational,okada2020planet,sekar2020planning} including Dreamer.
RSSM assumes the latent $\latent_{t}$ comprises $\latent_{t} = (\latents_{t}, \latenth_{t})$
where $\latents_{t}$, $\latenth_{t}$ are the probabilistic and deterministic variables, respectively.
RSSM's generative and inference models are defined as:
\begin{align}
  \mathrm{Generative\ models}&:
  \begin{cases}
    \latenth_{t} = f^{\mathrm{GRU}}(\latenth_{t-1}, \latents_{t-1}, \action_{t-1}) \\
    \latents_{t} \sim p(\latents_{t} | \latenth_{t}) \\
    \obs_{t} \sim p(\obs_{t} | \latenth_{t}, \latents_{t})
  \end{cases}, \\\nonumber
  \mathrm{Inference\ model}&: \latents_{t} \sim q(\latents_{t} | \latenth_{t}, \obs_{t}),
\end{align}
where deterministic $\latenth_{t}$ is considered to be the hidden state of the gated recurrent unit (GRU) $f^{\mathrm{GRU}}(\cdot)$~\cite{cho2014learning} so that historical information can be embedded into $\latenth_{t}$.

Dreamer~\cite{hafner2019dreamer} makes use of RSSM as a differentiable dynamics and efficiently learns the behaviors via backpropagation of Bellman errors estimated from imagined trajectories.
Dreamer's training procedure is simply summarized as follows:
\textit{(1)} Train RSSM with a given dataset by optimizing Eq.~(\ref{eqn:vae_elbo}).
\textit{(2)} Train a policy from the latent imaginations.
\textit{(3)} Execute the trained policy in a real environment and augment the dataset with the observed results.
The above steps are iteratively executed until the policy performs as expected.

\subsection{Contrastive Learning of RSSM} \label{sec:contrastive_rssm}
The original Dreamer paper~\cite{hafner2019dreamer} also introduced a likelihood-free objective
by reformulating $\mathcal{J}^{\mathrm{likelihood}}$ of Eq.~(\ref{eqn:vae_elbo}).
By adding a constant $\log p(\obs_{t})$ and applying Bayes' theorem, we get a decoder-free objective:
\begin{align}
  & \mathcal{J}^{\mathrm{likelihood}} \stackrel{+}{=} \mathbb{E}_{q(\latent_{t} | \cdot)} \left[
  \log p(\obs_{t}|\latent_{z}) - \log p(\obs_{t})
  \right] \nonumber \\
  & =
  \mathbb{E}_{q(\latent_{t} | \cdot)} \left[
  \log p(\latent_{t}|\obs_{t}) - \log p(\latent_{t})
  \right] \nonumber \\
  & \geq \mathbb{E}_{q(\latent_{t} | \cdot)} \left[
  \log p(\latent_{t}|\obs_{t}) - \log \sum_{\obs'\in \minibatch} p(\latent_{t}|\obs')
  \right] \nonumber \\
  & \coloneqq \nceloss, \label{eqn:cont1}
\end{align}
where $\minibatch$ denotes the mini-batch and the lower bound in the second line was from the Info-NCE (noise-contrastive estimator) mini-batch bound~\cite{poole2019variational}.
Let $B$ be the batch size of $\minibatch$, $\nceloss$ is considered as a $B$-class categorical cross entropy objective to discriminate the positive pair $(\latent_{t}, \obs_{t})$ among the other negative pairs $(\latent_{t}, \obs'(\neq\obs_{t}))$.
In this interpretation, $p(\latent_{t}|\obs_{t})$ can be considered as a discriminator to discern the positive pairs.
Representation learning with this type of objective is known as contrastive learning~\cite{oord2018representation,chen2020simple} that encourages the embeddings to be sufficiently seperated from each other in the latent space.
However, the experiment in \cite{hafner2019dreamer} has demonstrated that this objective significantly degrades the performance compared to the original objective $\mathcal{J}^{\mathrm{likelihood}}$.


\section{Proposed Contrastive Learning \\ and MBRL Method} \label{sec:proposed_method}
\subsection{Deriving Another Contrastive Objective} \label{sec:reformulation}
We propose to further reformulate $\nceloss$ of Eq.~(\ref{eqn:cont1}) by introducing a multi-step prediction model:
%
$
  \tilde{p}(\latent_{t} | \latent_{t-k}, \action_{<t})
  \coloneqq
  \mathbb{E}_{\tilde{p}(\latent_{t-1} | \latent_{t-k}, \action_{<t-1})}
  \left[
    \tilde{p}(\latent_{t}|\latent_{t-1},\action_{t-1})
  \right].
$
%
The accent of $\tilde{p}$ implies that an independent dynamics model from  $p(\latent_{t}|\latent_{t-1},\action_{t-1})$ in Eq.~(\ref{eqn:vae_elbo}) can be employed here.
By multiplying a constant $\mathbb{E}_{q(\latent_{t-k}|\cdot)}[\tilde{p}(\cdot)/\tilde{p}(\cdot)] = 1$, we obtain an importance sampling form of $\nceloss$ as:
\begin{align}
  \nceloss
  = &\ 
  \mathbb{E}_{\tilde{p}(\latent_{t}|\latent_{t-k},\action_{<t})q(\latent_{t-k}|\cdot)}\left[
    \frac{q(\latent_{t} | \cdot)}{\tilde{p}(\latent_{t}|\cdot)} \times \hookleftarrow
    \right.
    \nonumber \\ & \left.
    \left(
      \log p(\latent_{t}|\obs_{t}) - \log \sum_{\obs'} p(\latent_{t}|\obs')
    \right)
  \right]. \label{eqn:cont2}
\end{align}
For computational simplicity, we approximate the likelihood ratio ${q(\latent_{t} | \cdot)}/{\tilde{p}(\latent_{t}|\cdot)}$ as a constant and assume that the summation of $\nceloss$ across batch and time dimension is approximated as:
%
$ 
    \sum \nceloss \stackrel{\sim}{\propto} \sum \nceloss_{k},
$ 
where
\begin{align}
    & \nceloss_{k} \coloneqq \\
    & \mathbb{E}_{\tilde{p}(\latent_{t}|\latent_{t-k},\action_{<t})q(\latent_{t-k}|\cdot)}\left[
        \log p(\latent_{t}|\obs_{t}) - \log \sum_{\obs'} p(\latent_{t}|\obs')
    \right]. \nonumber
\end{align}
We further import the concept of \textit{overshooting} and optimize $\nceloss_{k}$ along with $\klloss_{k}$ on multi-step prediction of varying $k$s.
Finally, the objective we use to train RSSM is:
\begin{align}
  \mathcal{J} \coloneqq \textstyle\sum^{K}_{k=0} \left(\nceloss_{k} + \klloss_{k}\right). \label{eqn:proposed_objective}
\end{align}
%
Note that $\nceloss_{k}$ and $\klloss_{k}$ have different dynamics model (i.e., $\tilde{p}(\latent_{t}|\cdot)$ and $p(\latent_{t}|\cdot)$, respectively).

\subsection{Relation among the Objectives}
As shown in Appx.~\ref{sec:infomax}, $\nceloss$ is a lower bound of the mutual information $I(\obs_{t};\latent_{t})$,
while $\nceloss_{k}$ is a bound of $I(\obs_{t};\latent_{t-k})$.
Since the latent state sequence is Markovian, we have the data processing inequality as $I(\obs_{t};\latent_{t}) \geq I(\obs_{t};\latent_{t-k})$.
In other words, $\nceloss$ and approximately derived $\nceloss_{k}$ share the same InfoMax upper bound metrics.
An intuitive motivation to introduce $\nceloss_{k}$ instead of $\nceloss$ is so that we can incorporate temporal correlation between $t$ and $t-k$.
Another motivation is that we can increase the model capacity of the discriminator $p(\latent_{t}|\obs_{t})$ by incorporating the independent dynamics model $\tilde{p}(\latent_{t}|\cdot)$.

\subsection{Model Definitions} \label{sec:model_definition}
This section discusses how we define the discriminator components: $p(\latent_{t}|\obs_{t})$ and $\tilde{p}(\latent_{t}|\latent_{t-1},\action_{t-1})$. 
Ref.~\cite{tschannen2019mutual} has empirically shown that the inductive bias from model architectures is a significant factor for contrastive learning.
As experimentally recommended in the literature, we define $p(\latent_{t}|\obs_{t})$ as an exponentiated bilinear similarity function parameterized with $W_{\latent|\obs}$:
\begin{align}
  p(\latent_{t}|\obs_{t}) \propto \exp(\latent_{t}^{\top}W_{\latent|\obs}\boldsymbol{e}_{t}), \label{eqn:bilinear}
\end{align}
where $\boldsymbol{e}_{t} \coloneqq f^{\mathrm{CNN}}(\obs_{t})$ and $f^{\mathrm{CNN}}(\cdot)$ denotes feature extraction by a CNN unit.
With this definition, $\nceloss_{k}$ is simply a softmax cross-entropy objective with logits $\latent^{\top} W \boldsymbol{e}$. 
Contrary to the previous contrastive learning literature~\cite{oord2018representation,tschannen2019mutual},
the definition of newly introduced $\tilde{p}(\latent_{t}|\latent_{t-1},\action_{t-1})$ is required.
Here, we propose to apply linear modeling to define the model deterministically as:
\begin{align}
  & \tilde{p}(\latent_{t}|\latent_{t-1},\action_{t-1})
  \coloneqq
  \delta(\latent_{t} - \latent'_{t}),\ 
  \nonumber \\
  & \mathrm{where}\
  \latent'_{t} \coloneqq
  W_{\latent}\latent_{t-1} + W_{\action}\action_{t-1}, \label{eqn:linear_dynamics}
\end{align}
$\delta$ is the Dirac delta function, and $W_{\latent,\action}$ are linear parameters.

This linear modeling of $\tilde{p}(\latent_{t}|\cdot)$ successfully regularizes $\nceloss_{k}$ and contributes to construct smooth latent space.
We can alternatively define $\tilde{p}(\latent_{t}|\cdot) \coloneqq p(\latent_{t}|\cdot)$,
where $p(\latent_{t}|\cdot)$ is generally defined as an expressive model aiming at precise prediction.
However, the high model capacity allows to embed temporally consecutive samples too distant from each other to sufficiently optimize $\nceloss_{k}$,
thus yielding unsmooth latent space.

\subsection{Instantiation with RSSM} \label{sec:instantiation}
\begin{figure}
  \centering
  \includegraphics[width=0.48\textwidth]{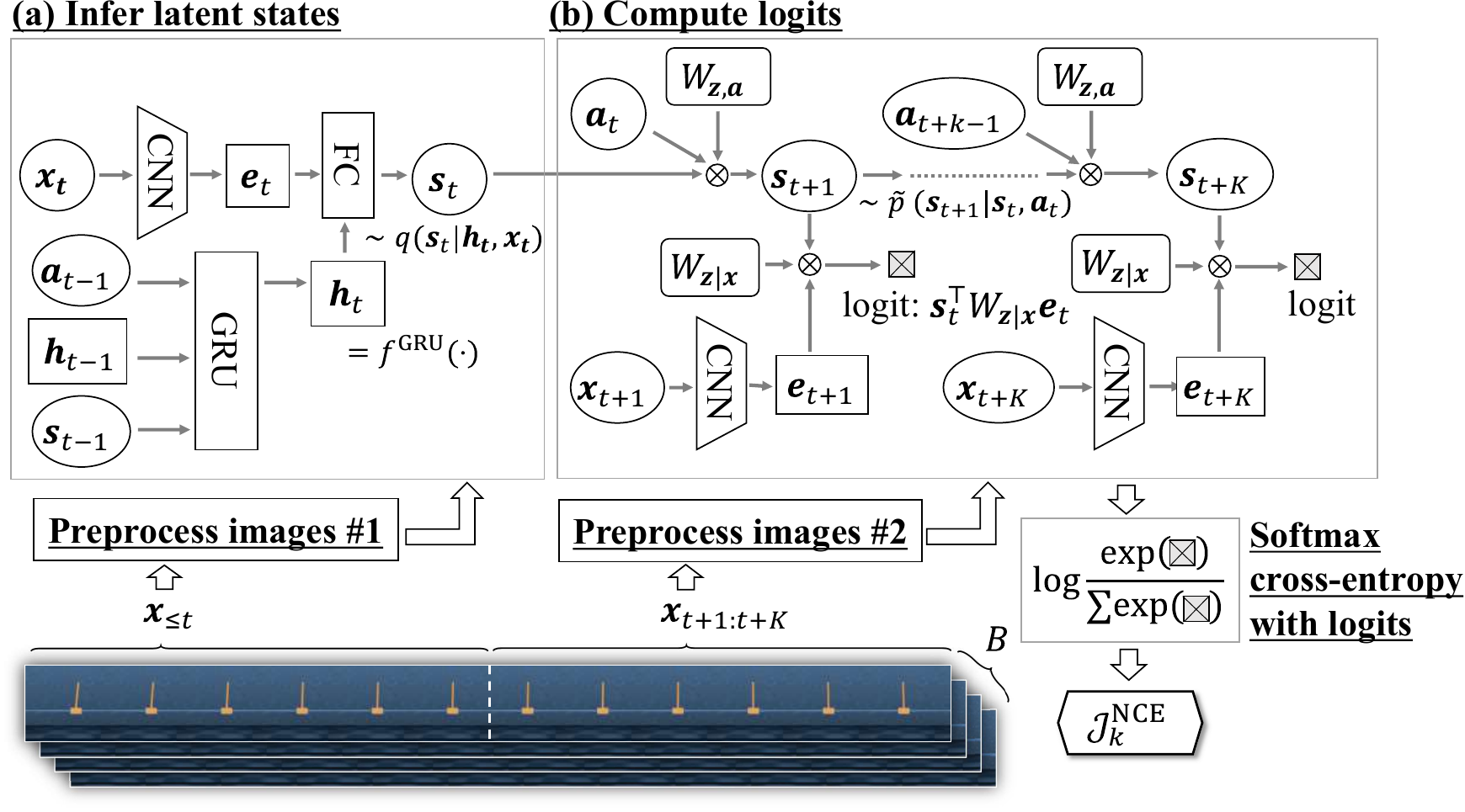}
  \caption{
   The RSSM-based architecture to compute $\nceloss_{k}$.
   \texttt{CNN}, \texttt{GRU} and \texttt{FC} represent a convolutional neural network, a GRU-cell, and a fully-connected layer, respectively.
   In module (a), latent states $\latents_{t}$ are recurrently inferred given $\obs_{\leq t}$.
   In module (b), $\latents_{t+1:t+K}$ are sequentially predicted by the linear model $W_{\latent,\action}$, and then they are compared with the observations $\obs_{t+1:t+K}$ to compute logits.
   For readability, we only illustrate positive logits,
   however, negative logits are also computed by pairing samples from different time-steps or frames,
   yielding $(B \times K)^{2}$ logits.
   To compute $\nceloss_{k}$ of a certain positive logit,
   remaining $(B \times K)^{2} - 1$ logits are used for negative logits.
  }
  \label{fig:rssm_cpc}
\end{figure}

Figure \ref{fig:rssm_cpc} illustrates the architecture to compute $\nceloss_{k}$.
We describe the two paramount components which characterize our proposed contrastive learning scheme as follows:

\textbf{(i) Independent linear forward dynamics:}
As previously proposed in Sec.~\ref{sec:model_definition}, we employ a simple linear forward dynamics $\tilde{p}$, which is used only for contrastive learning.
During the policy optimization phase, the expressive model with GRU is alternatively utilized
to make the most out of its long-term prediction accuracy.

\textbf{(ii) Data augmentation:}
We append two independent image preprocessors which process two sets of input images (i.e., $\obs_{\leq t}$ and $\obs_{t+1:t+K}$).
Considering the empirical success of the previous literature~\cite{chen2020simple,srinivas2020curl,laskin2020reinforcement,lee2020sunrise,kostrikov2020image}, we adopt the random crop of images.
In our implementation, the original image shaped $(72, 72)$ is cropped to be $(64, 64)$.
The origin of the crop rectangle is determined at each preprocessor randomly and indenpendently.
This makes it difficult for the contrastive learner to discriminate correct positive pairs, encouraging only informative features for control to be extracted.
%
%
%

We propose a decoder-free variant of Dreamer, which we call \textit{\underline{Dream}er with \underline{In}foMax and without \underline{g}enerative decoder} (\proposedmethod{}).
\proposedmethod{} trains a policy as almost same way with the original Dreamer. 
The only difference between the methods is that we alternatively use the contrastive learning scheme introduced in the previous section to train RSSM.
We implement \proposedmethod{} in TensorFlow~\cite{tensorflow2015-whitepaper} by modifying the official source code of Dreamer%
\footnote{\url{https://github.com/google-research/dreamer}}.
We keep all hyperparameters and experimental conditions similar to the original ones.
A newly introduced hyperparameter $K$ in Eq.~(\ref{eqn:proposed_objective}) (overshooting distance) is set to be $K=3$ based on the ablation study in Appx.~\ref{sec:ablation_os_distance}.

%
%
\section{Experiments} \label{sec:experiments}
\subsection{Comparison to State-of-the-art Methods}
\begin{table*}[t]
    \centering
    \caption{
    Performance on 15 benchmark tasks around 500K environment steps (100K only for Cup-catch).
    }
    \label{tab:comp_sota}
    \resizebox{0.75\textwidth}{!}{
    \begin{tabular}{lccc|cccc}
    \toprule
    & \multirow{2}{*}{\proposedmethod{} (ours)}
    & Dreamer w/
    & Dreamer w/
    & \multirow{2}{*}{CURL~\cite{srinivas2020curl}}
    & \multirow{2}{*}{DrQ~\cite{kostrikov2020image}}
    & \multirow{2}{*}{RAD~\cite{laskin2020reinforcement}} \\
    &
    & $\mathcal{J}^{\mathrm{likelihood}}$~\cite{hafner2019dreamer}
    & $\nceloss$ ~\cite{hafner2019dreamer}
    &
    &
    & \\
    \midrule
    \multicolumn{6}{l}{\textbf{(A) Manipulation tasks where \textit{object vanishing} is critical}}
    \\
    \midrule
    Cup-catch (100K)
    & {\begin{tabular}[c]{@{}c@{}} \best{925} $\pm$ \scriptsize{48} \end{tabular}}
    & {\begin{tabular}[c]{@{}c@{}} 698 $\pm$ \scriptsize{350} \end{tabular}}
    & {\begin{tabular}[c]{@{}c@{}} 609 $\pm$ \scriptsize{404} \end{tabular}}
    & {\begin{tabular}[c]{@{}c@{}} 693 $\pm$ \scriptsize{334} \end{tabular}}
    & {\begin{tabular}[c]{@{}c@{}} \second{882} $\pm$ \scriptsize{174} \end{tabular}}
    & {\begin{tabular}[c]{@{}c@{}} 792 $\pm$ \scriptsize{315} \end{tabular}}
    \\
    Reacher-hard
    & {\begin{tabular}[c]{@{}c@{}} \best{868} $\pm$ \scriptsize{272} \end{tabular}}
    & {\begin{tabular}[c]{@{}c@{}} 8 $\pm$ \scriptsize{33} \end{tabular}}
    & {\begin{tabular}[c]{@{}c@{}} 115 $\pm$ \scriptsize{298} \end{tabular}}
    & {\begin{tabular}[c]{@{}c@{}} 431 $\pm$ \scriptsize{435} \end{tabular}}
    & {\begin{tabular}[c]{@{}c@{}} 616 $\pm$ \scriptsize{464} \end{tabular}}
    & {\begin{tabular}[c]{@{}c@{}} 783 $\pm$ \scriptsize{370} \end{tabular}}
    \\
    Finger-turn-hard
    & {\begin{tabular}[c]{@{}c@{}} \best{752} $\pm$ \scriptsize{325} \end{tabular}}
    & {\begin{tabular}[c]{@{}c@{}} 264 $\pm$ \scriptsize{368} \end{tabular}}
    & {\begin{tabular}[c]{@{}c@{}} 222 $\pm$ \scriptsize{379}\end{tabular}}
    & {\begin{tabular}[c]{@{}c@{}} \second{339} $\pm$ \scriptsize{443} \end{tabular}}
    & {\begin{tabular}[c]{@{}c@{}} 270 $\pm$ \scriptsize{427} \end{tabular}}
    & {\begin{tabular}[c]{@{}c@{}} 303 $\pm$ \scriptsize{443} \end{tabular}}
    \\
    UR5-reach
    & {\begin{tabular}[c]{@{}c@{}} \best{845} $\pm$ \scriptsize{147} \end{tabular}}
    & {\begin{tabular}[c]{@{}c@{}} 652 $\pm$ \scriptsize{230} \end{tabular}}
    & {\begin{tabular}[c]{@{}c@{}} 592 $\pm$ \scriptsize{271} \end{tabular}}
    & {\begin{tabular}[c]{@{}c@{}} \second{729} $\pm$ \scriptsize{201} \end{tabular}}
    & {\begin{tabular}[c]{@{}c@{}} 633 $\pm$ \scriptsize{312} \end{tabular}}
    & {\begin{tabular}[c]{@{}c@{}} 642 $\pm$ \scriptsize{274} \end{tabular}}
    \\
    Connector-insert
    & {\begin{tabular}[c]{@{}c@{}} \best{629} $\pm$ \scriptsize{391} \end{tabular}}
    & {\begin{tabular}[c]{@{}c@{}} 169 $\pm$ \scriptsize{348} \end{tabular}}
    & {\begin{tabular}[c]{@{}c@{}} 304 $\pm$ \scriptsize{399} \end{tabular}}
    & {\begin{tabular}[c]{@{}c@{}} 297 $\pm$ \scriptsize{384} \end{tabular}}
    & {\begin{tabular}[c]{@{}c@{}} 183 $\pm$ \scriptsize{361} \end{tabular}}
    & {\begin{tabular}[c]{@{}c@{}} \second{367} $\pm$ \scriptsize{387} \end{tabular}}
    \\
    \midrule\midrule
    \multicolumn{6}{l}{(B) Manipulation tasks where \textit{object vanishing} is NOT critical}
    \\
    \midrule
    Reacher-easy
    & {\begin{tabular}[c]{@{}c@{}} \second{905} $\pm$ \scriptsize{210} \end{tabular}}
    & {\begin{tabular}[c]{@{}c@{}} \best{947} $\pm$ \scriptsize{145} \end{tabular}}
    & {\begin{tabular}[c]{@{}c@{}} 183 $\pm$ \scriptsize{325} \end{tabular}}
    & {\begin{tabular}[c]{@{}c@{}} 834 $\pm$ \scriptsize{286} \end{tabular}}
    & -
    & -
    \\
    Finger-turn-easy
    & {\begin{tabular}[c]{@{}c@{}} \second{661} $\pm$ \scriptsize{394} \end{tabular}}
    & {\begin{tabular}[c]{@{}c@{}} \best{689} $\pm$ \scriptsize{394} \end{tabular}}
    & {\begin{tabular}[c]{@{}c@{}} 232 $\pm$ \scriptsize{398} \end{tabular}}
    & {\begin{tabular}[c]{@{}c@{}} 576 $\pm$ \scriptsize{464} \end{tabular}}
    & -
    & -
    \\
    Finger-spin
    & {\begin{tabular}[c]{@{}c@{}} 762 $\pm$ \scriptsize{113} \end{tabular}}
    & {\begin{tabular}[c]{@{}c@{}} 763 $\pm$ \scriptsize{188} \end{tabular}}
    & {\begin{tabular}[c]{@{}c@{}} \second{886} $\pm$ \scriptsize{169} \end{tabular}}
    & {\begin{tabular}[c]{@{}c@{}} \best{922} $\pm$ \scriptsize{55} \end{tabular}}
    & -
    & -
    \\
    \midrule
    \multicolumn{6}{l}{(C) Pole-swingup tasks}
    \\
    \midrule
    Pendulum-swingup
    & {\begin{tabular}[c]{@{}c@{}} 811 $\pm$ \scriptsize{98} \end{tabular}}
    & {\begin{tabular}[c]{@{}c@{}} \second{432} $\pm$ \scriptsize{408} \end{tabular}}
    & {\begin{tabular}[c]{@{}c@{}} \best{825} $\pm$ \scriptsize{106} \end{tabular}}
    & {\begin{tabular}[c]{@{}c@{}} 46 $\pm$ \scriptsize{207} \end{tabular}}
    & -
    & -
    \\
    Acrobot-swingup
    & {\begin{tabular}[c]{@{}c@{}} \best{267} $\pm$ \scriptsize{177} \end{tabular}}
    & {\begin{tabular}[c]{@{}c@{}} \second{98} $\pm$ \scriptsize{119} \end{tabular}}
    & {\begin{tabular}[c]{@{}c@{}} 48 $\pm$ \scriptsize{54} \end{tabular}}
    & {\begin{tabular}[c]{@{}c@{}} 4 $\pm$ \scriptsize{14} \end{tabular}}
    & -
    & -
    \\
    Cartpole-swingup-sparse
    & {\begin{tabular}[c]{@{}c@{}} \best{465} $\pm$ \scriptsize{328} \end{tabular}}
    & {\begin{tabular}[c]{@{}c@{}} \second{317} $\pm$ \scriptsize{345} \end{tabular}}
    & {\begin{tabular}[c]{@{}c@{}} 197 $\pm$ \scriptsize{79} \end{tabular}}
    & {\begin{tabular}[c]{@{}c@{}} 17 $\pm$ \scriptsize{17} \end{tabular}}
    & -
    & -
    \\
    \midrule
    \multicolumn{6}{l}{(D) Locomotion tasks}
    \\
    \midrule
    Quadrupled-walk
    & {\begin{tabular}[c]{@{}c@{}} \best{719} $\pm$ \scriptsize{193} \end{tabular}}
    & {\begin{tabular}[c]{@{}c@{}} \second{441} $\pm$ \scriptsize{219} \end{tabular}}
    & {\begin{tabular}[c]{@{}c@{}} 201 $\pm$ \scriptsize{272} \end{tabular}}
    & {\begin{tabular}[c]{@{}c@{}} 188 $\pm$ \scriptsize{174} \end{tabular}}
    & -
    & -
    \\
    Walker-walk
    & {\begin{tabular}[c]{@{}c@{}} 469 $\pm$ \scriptsize{123} \end{tabular}}
    & {\begin{tabular}[c]{@{}c@{}} \best{955} $\pm$ \scriptsize{19} \end{tabular}}
    & {\begin{tabular}[c]{@{}c@{}} 483 $\pm$ \scriptsize{111} \end{tabular}}
    & {\begin{tabular}[c]{@{}c@{}} \second{914} $\pm$ \scriptsize{33} \end{tabular}}
    & -
    & -
    \\
    Cheetah-run
    & {\begin{tabular}[c]{@{}c@{}} 566 $\pm$ \scriptsize{118} \end{tabular}}
    & {\begin{tabular}[c]{@{}c@{}} \best{781} $\pm$ \scriptsize{132} \end{tabular}}
    & {\begin{tabular}[c]{@{}c@{}} 303 $\pm$ \scriptsize{174} \end{tabular}}
    & {\begin{tabular}[c]{@{}c@{}} \second{580} $\pm$ \scriptsize{56} \end{tabular}}
    & -
    & -
    \\
    Hopper-hop
    & {\begin{tabular}[c]{@{}c@{}} \second{78} $\pm$ \scriptsize{55} \end{tabular}}
    & {\begin{tabular}[c]{@{}c@{}} \best{172} $\pm$ \scriptsize{114} \end{tabular}}
    & {\begin{tabular}[c]{@{}c@{}} 25 $\pm$ \scriptsize{29} \end{tabular}}
    & {\begin{tabular}[c]{@{}c@{}} 10 $\pm$ \scriptsize{17} \end{tabular}}
    & -
    & -
    \\
    \bottomrule
    \end{tabular}}
\end{table*}

The main objective of this experiment is to demonstrate that \proposedmethod{} has advantages over the baseline method Dreamer~\cite{hafner2019dreamer}
on difficult 5 manipulation tasks exhibited in Fig.~\ref{fig:dazzling},
in which Dreamer suffers from \textit{object vanishing}.
We also prepare a likelihood-free variant of Dreamer introduced in Sec.~\ref{sec:contrastive_rssm},
which utilizes the vanilla contrastive objective $\nceloss$ instead of $\nceloss_{k}$ and $\mathcal{J}^{\mathrm{likelihood}}$.
The specifications of the two original tasks, UR5-reach and Connector-insert, are described in Appx.~\ref{sec:original_tasks}.
For the difficult 5 tasks, we also compare the performance with the latest cutting edge MFRL methods, which are CURL~\cite{srinivas2020curl}, DrQ~\cite{kostrikov2020image} and RAD~\cite{laskin2020reinforcement}.
In addition, another variety of 10 DMC tasks are evaluated, which are categorized into three classes namely; manipulation, pole-swingup, and locomotion.
For the additional 10 tasks, only CURL is selected as an MFRL representative.

Table~\ref{tab:comp_sota} summarizes the training results benchmarked at certain environment steps.
The results show the mean and standard deviation averaged 4 seeds and 10 consecutive trajectories.
This table shows a similar result as in \cite{hafner2019dreamer} that decoder-free Dreamer with the vanilla contrastive objective $\nceloss$ degrades the performances on most of tasks than decoder-based Dreamer with $\mathcal{J}^{\mathrm{likelihood}}$.
In the following discussions, we use the decoder-based Dreamer as a primary baseline.
(A) We put much focus on these difficult tasks and it can be seen that \proposedmethod{} consistently achieves better performance than Dreamer.
Hence, this indicates that the decoder-free nature of the proposed method successfully surmounts the \textit{object vanishing} problem.
In addition, \proposedmethod{} achieves outperforming performance than the leading MFRL methods.
(B) On other manipulation tasks, there are no significant difference between \proposedmethod{} and Dreamer because the key objects are large enough.
(C) Since the pole-swingup tasks also cause vanishing of thin poles,
\proposedmethod{} takes better performance than Dreamer.
(D) \proposedmethod{} lags behind the Dreamer on 3 of 4 locomotion tasks, i.e., planar locomotion tasks (Walker-walk, Cheetah-run and Hopper-hop).
On these tasks, the cameras always track the center of locomotive robots, and this causes the key control information (i.e., velocity) to be extracted from the background texture.
We suppose that this \textit{robot-centric} nature makes it difficult for the contrastive learner to extract such information
because only robots' attitudes provide enough information to discriminate different samples.

\begin{figure}[t]
  \centering
  \includegraphics[width=0.35\textwidth]{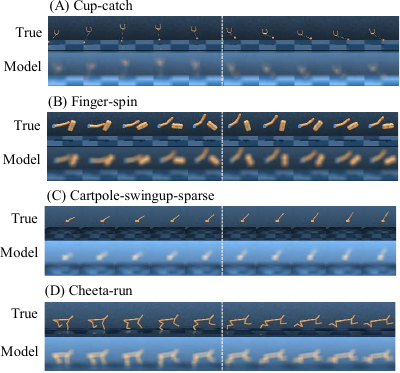}
  \caption{Open-loop video predictions.
   The left 5 consecutive images show reconstructed context frames and the remaining images are generated open-loop.
   The decoder is trained independently without backpropagating reconstruction errors to other models.
   }
  \label{fig:cpc_recons}
\end{figure}

Figure~\ref{fig:cpc_recons} shows video prediction by Dreaming,
in which principal features for control (e.g., positions and orientations) are successfully reconstructed from the embeddings learned without likelihood objective.
However on Cheeta-run, another kind of \textit{object vanishing} arises;
the checkered floor pattern, which is required to extract the velocity information, is vanished.
%
%
%

%

\subsection{Ablation Study}
\begin{table*}[t]
    \caption{
      Ablation study: The effects of $\mathrm{(i)}$ linear forward dynamics (left) and $\mathrm{(ii)}$ data augmentation (right).
    }
     \label{tab:ablation_linearity}
    \centering
    \resizebox{0.75\textwidth}{!}{
    \begin{tabular}{l|cc}
      \toprule
                            & \multicolumn{1}{c}{$\tilde{p}$: linear} 
                            & \multicolumn{1}{c}{$\tilde{p} \coloneqq p$} 
                            \\
                            & {as Eq.~(\ref{eqn:linear_dynamics})}
                            & as in \cite{yan2020learning}
                            \\
      \midrule
      Cup-catch \scriptsize{(100K)}
      & \best{925} $\pm$ \scriptsize{48}
      & \underline{575} $\pm$ \scriptsize{449}
      \\
      Reacher-hard &
       \best{868} $\pm$ \scriptsize{272}
      & \underline{232} $\pm$ \scriptsize{370}
      \\
      Finger-turn-hard &
       \best{752} $\pm$ \scriptsize{325}
       & \underline{263} $\pm$ \scriptsize{369}
       \\
      \bottomrule
    \end{tabular}
    \hspace{0.225cm}
    \begin{tabular}{l|c|ccc}
      \toprule
      Random crop  & - & $\checkmark$ & - & $\checkmark$ \\ 
      Color jitter & - & - & $\checkmark$ & $\checkmark$ \\ 
      \midrule
      Cup-catch \scriptsize{(100K)} & \underline{866} $\pm$ \scriptsize{133} & \best{925} $\pm$ \scriptsize{48} & 846 $\pm$ \scriptsize{192} & \underline{866} $\pm$ \scriptsize{121} \\ 
      Reacher-hard & 11 $\pm$ \scriptsize{32} & \best{868} $\pm$ \scriptsize{272} & 121 $\pm$ \scriptsize{292} & \second{733} $\pm$ \scriptsize{388}
      \\ 
      Finger-turn-hard & 114 $\pm$ \scriptsize{283} & \best{752} $\pm$ \scriptsize{325} & 191 $\pm$ \scriptsize{357} & \underline{399} $\pm$ \scriptsize{440} \\
      \bottomrule
    \end{tabular}
    }
\end{table*}

This experiment is conducted to analyze how the major components of the proposed representation learning, introduced in Sec.~\ref{sec:instantiation}, contribute to the overall performance.
For this purpose, some variants of the proposed method have been prepared:
{(i) the effect of independent linear dynamics} is demonstrated with a variant that has shared dynamics $\tilde{p}(\latent_{t}|\latent_{t-1}, \action_{t-1}) \coloneqq {p}(\latent_{t}|\latent_{t-1}, \action_{t-1})$%
\footnote{
The prepared variant can be considered as a special case of the contrastive forward model (CFM)~\cite{yan2020learning} as discussed in Sec.~\ref{sec:related_work}.
},
{(ii) the effect of data augmentation} is demonstrated by removing the image preprocessors shown in Fig.~\ref{fig:rssm_cpc}.
We also prepare another data augmentation called color jittering~\cite{chen2020simple,kostrikov2020image,laskin2020reinforcement}, for reference.
Only three tasks, which are Cup-catch, Reacher-hard, and Finger-turn-hard, are taken into this experiment.
Tables~\ref{tab:ablation_linearity} summarize the results from the performed ablation study,
which reveals that both of the proposed components are essential to achieve state-of-the-art results.
%

\section{Conclusion} \label{sec:conclusion}
In the present paper, we proposed \proposedmethod{}, a decoder-free extension of the state-of-the-art MBRL method from pixels, Dreamer.
A likelihood-free contrastive objective was derived by reformulating the original ELBO of Dreamer.
We incorporated the two indispensable components below to the contrastive learning:
(i) independent and linear forward dynamics, (ii) the random crop data augmentation.
By making the most of the decoder-free nature and the two components, \proposedmethod{} was able to outperform the baseline methods
on difficult tasks especially where Dreamer suffers from \textit{object vanishing}.

An disadvantage we observed in the experiments was that \proposedmethod{} degraded the training performance on planar locomotion tasks (e.g., Walker-walk),
where the contrastive learner has to focus on not only robots but also the background texture.
This weak point should be resolved in future work as it may affect industrial manipulation tasks where first-person-view from robots dynamically changes.
Another future research direction is to incorporate the uncertainty-aware concepts proposed in recent MBRL studies~\cite{chua2018deep,okada2019variational,okada2020planet,lee2020sunrise}.
Although we have achieved state-of-the-art results on some difficult tasks, we have often observed overfitted behaviors during the early training phase.
We believe that this \textit{model-bias} problem~\cite{deisenroth2011pilco} can be successfully solved by the above state-of-the-art strategy. 

\begin{appendices}

\section{Derivation} \label{sec:infomax}
In this section, we clarify that $\nceloss_{k}$ is a lower bound of $I(\obs_{t}, \latent_{t-k})$.
$\nceloss_{k}$ can be rewriten as:
\begin{align}
  & \nceloss_{k} = \nonumber \\ & \mathbb{E}_{q(\latent_{t-k}|\cdot)}\left[
  \log f(\obs_{t}, \latent_{t-k}) - \log\sum_{\obs' \in \mathcal{D}} f(\obs', \latent_{t-k})
  \right], \label{eqn:nceloss_rewrite}
\end{align}
where $f(\obs_{t}, \latent_{t-k})$ includes deterministic multi-step prediction with  $\tilde{p}(\latent_{t}|\latent_{t-k}, \action_{<t})$ and computation of the bilinear similarity by Eq.~(\ref{eqn:bilinear}).
For ease of notation, actions $\action_{<t}$ in the conditioning set are omitted from $f(\cdot)$.
As already shown in \cite{oord2018representation}, the optimal value of $f(\cdot)$ is given by:
\begin{align}
  f(\obs_{t}, \latent_{t-k}) \propto {p(\obs_{t}|\latent_{t-k})}/{p(\obs_{t})}.
\end{align}
By applying Bayes' theorem $f(\obs_{t}, \latent_{t-k}) \propto {p(\latent_{t-k}|\obs_{t})}/{p(\latent_{t-k})}$ and inserting this to Eq.~(\ref{eqn:nceloss_rewrite}), we get:
\begin{align}
  \nceloss_{k} &
  %
  \propto \mathbb{E}_{q(\latent_{t-k}|\cdot)}\left[
  \log {p(\latent_{t-k}|\obs_{t})} - \log\sum_{\obs'} {p(\latent_{t-k}|\obs')}
  \right] \nonumber\\
  &\leq \mathbb{E}_{q(\latent_{t-k}|\cdot)}\left[
  \log {p(\latent_{t-k}|\obs_{t})} - \log {p(\latent_{t-k})}
  \right]. \label{eqn:nceloss_mi}
\end{align}
By marginalizing Eq.~(\ref{eqn:nceloss_mi}) with respect to the data distribution, we finally obtain:
$ 
  \mathbb{E}[\nceloss_{k}] \leq I(\obs_{t}; \latent_{t-k}).
$ 
Note that setting $k=0$ derives $\mathbb{E}[\nceloss] \leq I(\obs_{t}; \latent_{t}).$

\section{Specifications of the Original Tasks} \label{sec:original_tasks}
\begin{figure}[tb]
  \centering
  \includegraphics[width=0.2\textwidth]{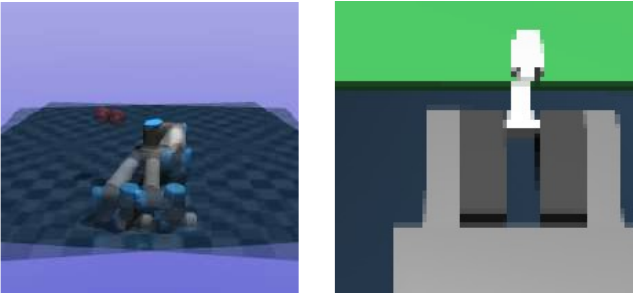}
  \caption{
    \textbf{UR5-reach} (left) is to bring the robot end effector
  to goal positions. The observation is a blended image of two different views, implicitly providing depth information.
  \textbf{Connector-insert} (right) is to insert a millimeter-sized connector gripped by a robot to a socket.
  This tasks is originally introduced in \cite{okumura2020domain}.
  Since the gap between the connector and socket is very tight, pixel-wise precise control is required.
  In the both tasks, the goal positions are initialized at random.
  }
  \label{fig:new_tasks}
\end{figure}

Figure \ref{fig:new_tasks} exhibits the specifications of newly introduced robotics tasks.

\section{Ablation Study of Overshooting Distance} \label{sec:ablation_os_distance}
Table \ref{tab:ablation_os} summarizes the ablation study of the overshooting distance $K$, 
which demonstrates that incorporating temporal correlation of appropriate multi-steps ($K=3$) is effective.
\begin{table}[tb]
  \caption{
  Ablation study: The effect of the overshooting distance $K$.
  }
  \label{tab:ablation_os}
  \centering
  \resizebox{0.45\textwidth}{!}{
  \begin{tabular}{l|cccc}
    \toprule
    & $K=1$ & $K=3$ & $K=5$ & $K=7$ \\ 
    \midrule
    Cup-catch \scriptsize{(100K)} & 280 $\pm$ \scriptsize{437} & \best{925} $\pm$ \scriptsize{48} & 734 $\pm$ \scriptsize{378} & \second{736} $\pm$ \scriptsize{378} \\ 
    Reacher-hard & 234 $\pm$ \scriptsize{364} & \best{868} $\pm$ \scriptsize{272} & \second{561} $\pm$ \scriptsize{447} & 471 $\pm$ \scriptsize{433}
    \\ 
    Finger-turn-hard & 354 $\pm$ \scriptsize{438} & \best{752} $\pm$ \scriptsize{325} & 468 $\pm$ \scriptsize{432} & \second{715} $\pm$ \scriptsize{375}
    \\ 
    \bottomrule
  \end{tabular}
  }
\end{table}

\end{appendices}

\clearpage
\bibliography{icra}

\begin{thebibliography}{10}

\bibitem{chua2018deep}
K.~Chua, R.~Calandra, R.~McAllister, and S.~Levine, ``Deep reinforcement
  learning in a handful of trials using probabilistic dynamics models,'' in
  {\em NeurIPS}, 2018.

\bibitem{okada2019variational}
M.~Okada and T.~Taniguchi, ``Variational inference {MPC} for {Bayesian}
  model-based reinforcement learning,'' in {\em CoRL}, 2019.

\bibitem{langlois2019benchmarking}
E.~Langlois, S.~Zhang, G.~Zhang, P.~Abbeel, and J.~Ba, ``Benchmarking
  model-based reinforcement learning,'' {\em arXiv:1907.02057}, 2019.

\bibitem{kaiser2019model}
L.~Kaiser, M.~Babaeizadeh, P.~Milos, B.~Osinski, {\em et~al.}, ``Model-based
  reinforcement learning for {Atari},'' in {\em ICLR}, 2020.

\bibitem{haarnoja2018soft}
T.~Haarnoja, A.~Zhou, P.~Abbeel, and S.~Levine, ``Soft actor-critic: Off-policy
  maximum entropy deep reinforcement learning with a stochastic actor,'' in
  {\em ICML}, 2018.

\bibitem{bechtle2020curious}
S.~Bechtle, Y.~Lin, A.~Rai, L.~Righetti, and F.~Meier, ``Curious {iLQR}:
  Resolving uncertainty in model-based {RL},'' in {\em CoRL}, 2019.

\bibitem{nagabandi2020deep}
A.~Nagabandi, K.~Konolige, S.~Levine, and V.~Kumar, ``Deep dynamics models for
  learning dexterous manipulation,'' in {\em CoRL}, 2019.

\bibitem{yang2020data}
Y.~Yang, K.~Caluwaerts, A.~Iscen, T.~Zhang, J.~Tan, and V.~Sindhwani, ``Data
  efficient reinforcement learning for legged robots,'' in {\em CoRL}, 2019.

\bibitem{zhang2019asynchronous}
Y.~Zhang, I.~Clavera, B.~Tsai, and P.~Abbeel, ``Asynchronous methods for
  model-based reinforcement learning,'' in {\em CoRL}, 2019.

\bibitem{williams2020locally}
G.~R. Williams, B.~Goldfain, K.~Lee, J.~Gibson, J.~M. Rehg, and E.~A.
  Theodorou, ``Locally weighted regression pseudo-rehearsal for adaptive model
  predictive control,'' in {\em CoRL}, 2019.

\bibitem{fang2019dynamics}
K.~Fang, Y.~Zhu, A.~Garg, S.~Savarese, and L.~Fei-Fei, ``Dynamics learning with
  cascaded variational inference for multi-step manipulation,'' in {\em CoRL},
  2019.

\bibitem{kingma2013auto}
D.~P. Kingma and M.~Welling, ``Auto-encoding variational bayes,'' in {\em
  ICLR}, 2014.

\bibitem{ha2018recurrent}
D.~Ha and J.~Schmidhuber, ``Recurrent world models facilitate policy
  evolution,'' in {\em NeurIPS}, 2018.

\bibitem{lee2019stochastic}
A.~X. Lee, A.~Nagabandi, P.~Abbeel, and S.~Levine, ``Stochastic latent
  actor-critic: Deep reinforcement learning with a latent variable model,''
  {\em arXiv:1907.00953}, 2019.

\bibitem{hafner2018learning}
D.~Hafner, T.~Lillicrap, I.~Fischer, R.~Villegas, D.~Ha, H.~Lee, and
  J.~Davidson, ``Learning latent dynamics for planning from pixels,'' in {\em
  ICML}, 2019.

\bibitem{hafner2019dreamer}
D.~Hafner, T.~Lillicrap, J.~Ba, and M.~Norouzi, ``Dream to control: Learning
  behaviors by latent imagination,'' {\em ICLR}, 2020.

\bibitem{han2019variational}
D.~Han, K.~Doya, and J.~Tani, ``Variational recurrent models for solving
  partially observable control tasks,'' in {\em ICLR}, 2020.

\bibitem{yarats2019improving}
D.~Yarats, A.~Zhang, I.~Kostrikov, B.~Amos, J.~Pineau, and R.~Fergus,
  ``Improving sample efficiency in model-free reinforcement learning from
  images,'' {\em arXiv:1910.01741}, 2019.

\bibitem{okada2020planet}
M.~Okada, N.~Kosaka, and T.~Taniguchi, ``{PlaNet} of the {Bayesians}:
  Reconsidering and improving deep planning network by incorporating {Bayesian}
  inference,'' in {\em IROS}, 2020.

\bibitem{deepmindcontrolsuite2018}
Y.~Tassa, Y.~Doron, A.~Muldal, T.~Erez, Y.~Li, {\em et~al.}, ``Deep{Mind}
  control suite,'' {\em arXiv:1801.00690}, 2018.

\bibitem{yan2020learning}
W.~Yan, A.~Vangipuram, P.~Abbeel, and L.~Pinto, ``Learning predictive
  representations for deformable objects using contrastive estimation,'' {\em
  arXiv:2003.05436}, 2020.

\bibitem{oord2018representation}
A.~v.~d. Oord, Y.~Li, and O.~Vinyals, ``Representation learning with
  contrastive predictive coding,'' {\em arXiv:1807.03748}, 2018.

\bibitem{chen2020simple}
T.~Chen, S.~Kornblith, M.~Norouzi, and G.~Hinton, ``A simple framework for
  contrastive learning of visual representations,'' in {\em ICLR}, 2020.

\bibitem{bellemare2013arcade}
M.~G. Bellemare, Y.~Naddaf, J.~Veness, and M.~Bowling, ``The arcade learning
  environment: An evaluation platform for general agents,'' {\em Journal of
  Artificial Intelligence Research}, vol.~47, pp.~253--279, 2013.

\bibitem{srinivas2020curl}
A.~Srinivas, M.~Laskin, and P.~Abbeel, ``{CURL}: Contrastive unsupervised
  representations for reinforcement learning,'' in {\em ICML}, 2020.

\bibitem{zhang2020learning}
A.~Zhang, R.~McAllister, R.~Calandra, Y.~Gal, and S.~Levine, ``Learning
  invariant representations for reinforcement learning without
  reconstruction,'' {\em arXiv:2006.10742}, 2020.

\bibitem{ma2020discriminative}
X.~Ma, P.~Karkus, D.~Hsu, W.~S. Lee, and N.~Ye, ``Discriminative particle
  filter reinforcement learning for complex partial observations,'' in {\em
  ICLR}, 2020.

\bibitem{laskin2020reinforcement}
M.~Laskin, K.~Lee, A.~Stooke, L.~Pinto, P.~Abbeel, and A.~Srinivas,
  ``Reinforcement learning with augmented data,'' {\em arXiv:2004.14990}, 2020.

\bibitem{kostrikov2020image}
I.~Kostrikov, D.~Yarats, and R.~Fergus, ``Image augmentation is all you need:
  Regularizing deep reinforcement learning from pixels,'' {\em
  arXiv:2004.13649}, 2020.

\bibitem{lee2020sunrise}
K.~Lee, M.~Laskin, A.~Srinivas, and P.~Abbeel, ``{SUNRISE}: A simple unified
  framework for ensemble learning in deep reinforcement learning,'' {\em
  arXiv:2007.04938}, 2020.

\bibitem{sekar2020planning}
R.~Sekar, O.~Rybkin, K.~Daniilidis, P.~Abbeel, D.~Hafner, and D.~Pathak,
  ``Planning to explore via self-supervised world models,'' {\em
  arXiv:2005.05960}, 2020.

\bibitem{cho2014learning}
K.~Cho, B.~Van~Merri{\"e}nboer, C.~Gulcehre, D.~Bahdanau, {\em et~al.},
  ``Learning phrase representations using {RNN} encoder-decoder for statistical
  machine translation,'' {\em arXiv:1406.1078}, 2014.

\bibitem{poole2019variational}
B.~Poole, S.~Ozair, A.~v.~d. Oord, A.~A. Alemi, and G.~Tucker, ``On variational
  bounds of mutual information,'' in {\em ICML}, 2019.

\bibitem{tschannen2019mutual}
M.~Tschannen, J.~Djolonga, P.~K. Rubenstein, S.~Gelly, and M.~Lucic, ``On
  mutual information maximization for representation learning,'' {\em ICLR},
  2020.

\bibitem{tensorflow2015-whitepaper}
M.~Abadi, A.~Agarwal, P.~Barham, E.~Brevdo, {\em et~al.}, ``{TensorFlow}:
  Large-scale machine learning on heterogeneous systems,'' 2015.

\bibitem{deisenroth2011pilco}
M.~Deisenroth and C.~E. Rasmussen, ``{PILCO}: A model-based and data-efficient
  approach to policy search,'' in {\em ICML}, 2011.

\bibitem{okumura2020domain}
R.~Okumura, M.~Okada, and T.~Taniguchi, ``Domain-adversarial and -conditional
  state space model for imitation learning,'' in {\em IROS}, 2020.

\end{thebibliography}
\bibliographystyle{ieeetr}

\end{document}